\documentclass[aps,amsmath,amssymb,showkeys,12pt,tightenlines,nobibnotes,nofootinbib]{revtex4}%for JSTAT
\usepackage{amsfonts}
\usepackage{mathrsfs}
\usepackage{graphicx}
\usepackage{amsmath}
\usepackage{amssymb}
\usepackage{amsbsy}
\usepackage{bm}

\def\bs{\bm{\sigma}}
\def\bx{\bm{\xi}}
\def\bxh{\hat{\bm{\xi}}}
\def\hb{\tilde{\beta}}
\def\hG{\tilde{G}}

\begin{document}

\title{Statistical mechanics of unsupervised feature learning in a restricted Boltzmann machine with binary synapses}

\author{Haiping Huang}
\email{physhuang@gmail.com}
\affiliation{RIKEN Brain Science Institute, Wako-shi, Saitama
351-0198, Japan}
\date{\today}

\begin{abstract}
Revealing hidden features in unlabeled data is called unsupervised feature learning, which plays an important role in
pretraining a deep neural network. Here we provide a statistical mechanics analysis of the unsupervised learning in a restricted Boltzmann machine with binary synapses.
A message passing equation to infer the hidden feature is derived, and furthermore, variants of this equation are analyzed. 
A statistical analysis by replica theory describes the thermodynamic
properties of the model. Our analysis confirms an entropy crisis preceding the non-convergence of the
message passing equation, suggesting a discontinuous phase transition as a key characteristic of the restricted Boltzmann machine.
Continuous phase transition is also confirmed depending on the embedded feature strength in the data. The mean-field
result under the replica symmetric assumption agrees with that obtained by running message passing algorithms on single instances of finite
sizes. Interestingly, in an approximate Hopfield model, the entropy crisis is absent, and a continuous phase transition is observed instead. We also develop an
iterative equation to infer the hyper-parameter (temperature) hidden in the data, which in physics corresponds to iteratively imposing Nishimori condition.
Our study provides insights towards understanding the thermodynamic properties of the restricted Boltzmann
machine learning, and moreover important theoretical basis to build simplified deep networks.

\end{abstract}

%\pacs{02.50.Tt, 87.19.L-, 75.10.Nr}
%for JSTAT
 \maketitle
\section{Introduction}
%%%%%%%%%%%%%%%%%%%%%%%%%%%%%%%%%%%%%%%%%%%%%%%%%%%%%%%%%%%%%%%%%
Standard machine learning algorithms require
a huge number of training examples to uncover hidden features, while humans and other animals can learn new concepts from 
only a few examples without any supervision signal~\cite{Lake-2015}. Learning hidden features in unlabeled training examples is called
unsupervised learning. Understanding how the number of examples confines the learning process is of fundamental importance
in both cognitive neuroscience and machine learning~\cite{Kersten-2004,Hinton-2007}. As already observed in training of deep neural networks,
unsupervised pretraining can significantly enhance the final performance, because the unsupervised pretraining provides a good initial region 
in parameter space from which the 
final fine-tuning starts~\cite{Bengio-2013}. However, there are few theoretical works addressing how unsupervised learning extracts hidden features. 
One potential reason is that the unsupervised learning process in a deep neural network is typically very complicated. 
Hence, understanding the mechanism of unsupervised learning in simple models is of significant importance.

This topic is recently studied based on the Bayesian inference framework~\cite{Huang-2016}.
In this recent work, the authors treated each example (data) as a constraint on the factor graph, and reformulated the learning of model parameters as a 
Bayesian inference problem on graphical models, and derived the message passing equations to infer the hidden feature from a finite amount of data. They observed an entropy crisis in
a simple restricted Boltzmann machine (RBM) model,
and predicted a discontinuous phase transition. However, in an approximate Hopfield model 
obtained by
a high-temperature expansion of the RBM model, the entropy crisis is absent, and instead, a continuous transition is observed. These properties observed in studies of 
single instances capture
key characteristics of the unsupervised feature learning. 

Here, we further demonstrate that the message passing equation derived in the recent work~\cite{Huang-2016} agrees with the statistical analysis of
the system in the thermodynamic limit via replica theory, a standard theoretical tool in spin glass theory of disordered systems~\cite{MM-1987}. 
The replica computation predicts the location of phase transition separating an
impossible-to-infer regime to inferable regime. This transition can be continuous depending on the embedded
feature strength. A discontinuous phase transition always exists in a restricted Boltzmann machine learning, but absent in 
an approximate Hopfield model where only continuous phase transition is observed. We also develop an iterative equation to infer the hyper-parameter
(temperature) hidden in the data, which in physics corresponds to iteratively imposing Nishimori
condition. This iterative scheme can even quantitatively predict how apparent features embedded in a real dataset are.
Our analysis gives a thorough understanding of
novel properties of the restricted Boltzmann machine within replica symmetric approximation. 

This paper is structured as follows. In Sec.~\ref{srbmm}, we introduce a simple RBM model for unsupervised feature learning, and 
propose the Bayesian inference framework to derive the message passing equation on factor graph representation of the learning process,
and this equation is then statistically analyzed and compared with replica computation under the replica symmetric assumption. A more efficient
approximate message passing equation is also derived. We also derive an iterative equation based on Bayes rule to predict the unknown temperature (feature strength) in the data. In Sec.~\ref{ahopf}, we approximate the RBM with the Hopfield model where
the stored pattern is interpreted as the feature vector. Similar statistical analysis is carried out, and its physical implications are
discussed. We end the paper with a summary in Sec.~\ref{conc}.

\section{Simple restricted Boltzmann machine learning and its statistical mechanics properties}
\label{srbmm}
\subsection{Simple restricted Boltzmann machine learning with binary synapses}
\label{srbmbs}
Restricted Boltzmann machine is a basic unit widely used in building a deep belief network~\cite{Hinton-2006a,Bengio-2013}. It consists of two layers of neurons.
The visible layer receives the input examples while the other hidden layer builds an internal representation of the input. No lateral connections exist within each
layer for computational efficiency. The symmetric connections (synapses) between visible and hidden neurons are considered as features the network tries to learn from a large
number of training examples. 

It is a common strategy to use sampling-based gradient-decent method to learn features in the data~\cite{Bengio-2013}, however, the gradient-decent learning is complicated and not amenable for analytical studies. Recent work showed that learning features can also be studied within a Bayesian learning framework~\cite{Huang-2016}, which
has the advantage of accounting for the uncertainty (about the features) caused by noises in the data~\cite{wunc-2015}, and furthermore can be analytically studied on probabilistic graphical models. 

We focus on an unsupervised learning of finite samplings generated by a simple RBM,
where a single hidden neuron is considered. The task is to
uncover an unknown rule embedded in the unlabeled data. The rule is represented by a binary feature vector defined as $\{\xi_{i}\}$ where $i$ goes from $1$ to $N$, the number of neurons
in the visible layer. We assume components of the true hidden feature vector connecting the visible neurons and the hidden neuron can take only two values, i.e., $+1$ or $-1$, with equal probabilities. Then this
feature vector is used to generate independent random samples according to the joint probability $P(\bs,h)\propto e^{-\beta E(\bs,h)/\sqrt{N}}$, where $E(\bs,h)=-\sum_{i}h\xi_i\sigma_i$, and $\bs$ is the
visible configuration and $h$ is the hidden neuron's state. Both $h$ and components of $\bs$ take binary values ($\pm1$) as well.
$\xi_i$ represents the connection between visible neuron $\sigma_i$ and the unique hidden neuron $h$. Moreover, we assume $\bx$ is a binary feature vector in the current setting of unsupervised learning.
A rescaled feature factor by the system size is assumed as well.
The feature vector is also multiplied
by an inverse-temperature parameter $\beta$ to investigate effects of the feature strength on the unsupervised learning. For simplicity, we consider the case of neurons without any
external biases (fields). Generalization to the case of neurons with
external fields is straightforward (Appendix~\ref{gener}).

The distribution of $\bs$ can be obtained by marginalization of $h$ on the joint distribution $P(\bs,h)$, resulting in
\begin{equation}
 P(\bs|\bx)=\frac{\cosh\left(\frac{\beta}{\sqrt{N}}\bx^{{\rm T}}\bs\right)}{\sum_{\bs}\cosh\left(\frac{\beta}{\sqrt{N}}\bx^{{\rm T}}\bs\right)},
\end{equation}
where the normalization is in fact independent of $\bx$, since $\sum_{\bs}\cosh\left(\frac{\beta}{\sqrt{N}}\bx^{{\rm T}}\bs\right)=\Bigl[2\cosh\frac{\beta}{\sqrt{N}}\Bigr]^N$. 
Suppose we have $M$ independent samples or examples $\{\bs^{a}\}_{a=1}^{M}$ to learn
the true hidden feature vector $\bx$, using the Bayes' formula, we have the posterior distribution of the feature vector as
\begin{equation}\label{Pobs}
 P(\bx|\{\bs^{a}\})=\frac{\prod_{a}P(\bs^{a}|\bx)}{\sum_{\bx}\prod_{a}P(\bs^{a}|\bx)}=\frac{1}{Z}\prod_{a}\cosh\left(\frac{\beta}{\sqrt{N}}\bx^{{\rm T}}\bs^{a}\right),
\end{equation}
where $Z$ is the partition function of the model, $a$ goes over all examples and ${\rm T}$ denotes a vector transpose operation. A uniform prior probability for the feature vector is assumed for simplicity.
A large $\beta$ indicates the feature in the data is strong, and expected to be revealed by a few examples, while
a weak feature vector may not be revealed by a huge number of examples. Each example serves as a constraint
to the learning process. Once $M>1$, the model becomes non-trivial as the partition function could not be computed exactly for a large number of visible neurons. This $M$ can be finite or proportional to the system size, and in the latter case we define
a data density as $\alpha=M/N$. Hereafter, we omit the conditional dependence of $P(\bx|\{\bs^{a}\})$ on $\{\bs^{a}\}$.

\begin{figure}
\centering
    \includegraphics[bb=46 527 418 736,scale=0.55]{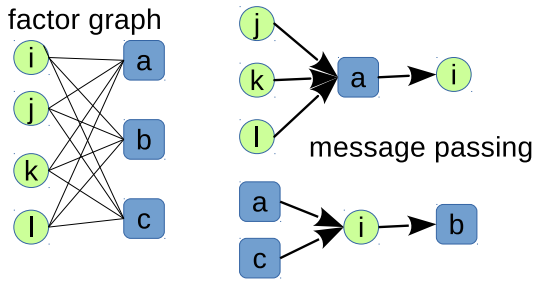}
  \caption{ (Color online) Schematic illustration of factor graph representation and message passing. Left panel: circle nodes indicate feature components to be
  inferred. Square nodes indicate data constraints. The strength of each connection is given by the data, e.g., $\sigma_i^{a}$ indicates the strength ($\pm1$) with which the feature component $\xi_i$ is related to $a$-th example.
  Right panel: the top panel shows constraint $a$ collects information from its neighboring feature nodes other than $i$ and produces an output message to node $i$. The bottom panel shows
  node $i$ collects information from its neighboring constraints other than $b$ and produces an output message to node $b$. The figure is taken from Ref~\cite{Huang-2016}.
  }\label{rbm}
\end{figure}
\subsection{Bayesian learning via message passing}
\label{blmp}
We call optimizing the marginal posterior probability of feature vectors given the data as Bayesian learning in the current unsupervised learning context. That is,
we compute the maximizer of the posterior marginals (MPM) estimator $\hat\xi_i = \arg\max_{\xi_i}P_i(\xi_i)$~\cite{Nishimori-2001}.
We define the overlap between the inferred feature vector and the true one as $q=\frac{1}{N}\sum_i\xi_i^{{\rm true}}\hat{\xi}_i$, where $\bxh$ is the inferred
 feature vector. The MPM estimator maximizes the overlap. If $q=0$, the examples do not give any information about the feature vector. If $q=1$, the feature vector is perfectly determined.
 In the numerical simulations, $q$ is evaluated by an average over many random instances (different true feature vectors). To compare with replica computation (Sec.~\ref{replicasrbm}), the average value $q=\left<\frac{1}{N}\sum_i\xi_i^{{\rm true}}\Bigl<\hat{\xi}_i\Bigr>\right>$ is used. Note that
 the inner average is the thermal average, and
 the outer average is taken with respect to different choices of true features.

 The statistical inference now
is simplified to the computation of marginal probabilities, e.g., $P_i(\xi_i)$, which is still a hard problem due to the interaction among example constraints. However, by mapping
the model (Eq.~(\ref{Pobs})) onto a factor graph~\cite{MM-2009,Huang-2015b}, the marginal probability can be estimated by message passing (Fig.~\ref{rbm}). 
For simplicity, we give the final simplified message passing equations (sMP) as follows (detailed derivations are given in Ref~\cite{Huang-2016}, also in Appendix~\ref{sMP-rbm}):
\begin{subequations}\label{bp0}
\begin{align}
m_{i\rightarrow a}&=\tanh\left(\sum_{b\in\partial i\backslash
a}u_{b\rightarrow i}\right),\\
u_{b\rightarrow i}&=\tanh^{-1}\left(\tanh(\beta G_{b\rightarrow i})\tanh(\beta\sigma_i^{b}/\sqrt{N})\right),\label{bp4}
\end{align}
\end{subequations}
where $G_{b\rightarrow
i}=\frac{1}{\sqrt{N}}\sum_{j\in\partial b\backslash i}\sigma_{j}^{b}m_{j\rightarrow b}$. The cavity magnetization is defined as $m_{j\rightarrow b}=\sum_{\xi_j}\xi_jP_{j\rightarrow b}(\xi_j)$. $m_{i\rightarrow a}$
can be interpreted as the message passing from feature $i$ to the data constraint $a$, while $u_{b\rightarrow i}$ can be
interpreted as the message passing from data constraint $b$ to its feature $i$. If the weak correlation assumption (also named Bethe approximation~\cite{cavity-2001}) is
self-consistent, the sMP would converge to a fixed point corresponding to a stationary point of the Bethe free energy function with respect to the cavity messages $\{m_{i\rightarrow a},u_{a\rightarrow i}\}$
~\cite{MM-2009}. From this fixed point, one can extract useful information about the true feature vector of the data, by calculating the marginal probability as
$P_i(\xi_i)=\frac{1+m_i\xi_i}{2}$ where $m_i=\tanh\left(\sum_{b\in\partial i}u_{b\rightarrow i}\right)$. Note that
we perform the feature inference using the same inverse-temperature as used to generate the data, thus the inference
is Bayes-optimal and satisfies the Nishimori condition~\cite{Nishimori-2001}. In the simulation section, we also perform the Bayesian inference using slightly different
temperatures.

Next, we compute the Bethe free energy, which is obtained by $-\beta Nf_{{\rm RS}}=\sum_{i}\ln Z_i-(N-1)\sum_a\ln Z_a$.
The free energy contribution of a feature node reads
\begin{equation}\label{Zirbm}
 \ln Z_i=\sum_{a\in\partial i}\left[\beta^2\Xi^{2}_{a\rightarrow i}/2+
\ln\cosh\Bigl(\beta G_{a\rightarrow i}+\beta\sigma_i^{a}/\sqrt{N}\Bigr)\right]+\ln\Bigl(1+\prod_{a\in\partial i}\mathcal{G}_{a\rightarrow i}\Bigr),
\end{equation}
and the free energy contribution of a data node reads
\begin{equation}\label{Zarbm}
\ln Z_a=\beta^2\Xi^2_{a}/2+\ln\cosh\beta G_a,
\end{equation}
where we define $\mathcal{G}_{a\rightarrow i}=e^{-2u_{a\rightarrow i}}$, $\Xi^{2}_{a\rightarrow i}\simeq\frac{1}{N}\sum_{j\in\partial a\backslash
i}(1-m_{j\rightarrow a}^{2})$, $G_a=\frac{1}{\sqrt{N}}\sum_{i\in\partial a}\sigma_i^{a}m_{i\rightarrow a}$, and $\Xi^2_a=\frac{1}{N}\sum_{i\in\partial a}(1-m^2_{i\rightarrow a})$.
%%%%%%%%%%%%%%%%%%%%%%%%%%%%%%%%%%%%%%%%%%%%%%%%%%%%%%%%%%%%%%%%%%%

Another important quantity is the number of feature vectors consistent with the presented random samplings, characterized by the entropy per neuron
$s=-\frac{1}{N}\sum_{\bx}P(\bx)\ln P(\bx)$. In the presence of a larger dataset, the generative machine should have less uncertainty about the underlying feature, corresponding to
small or vanishing entropy.
The entropy can be derived by using the 
standard thermodynamic formula $s=(1-\beta\frac{\partial}{\partial
\beta})(-\beta f_{{\rm RS}})$. Under the Bethe approximation, $s$ is evaluated as summing up contributions from single feature nodes and example nodes:
$Ns=\sum_i\Delta S_i-(N-1)\sum_a\Delta S_{a}$, where single feature node contribution is expressed as
\begin{equation}\label{Sirbm}
\begin{split}
\Delta S_i=\sum_{a\in\partial i}\left[\beta^2\Xi_{a\rightarrow i}^2/2+
\ln\cosh(\beta G_{a\rightarrow i}+\beta\sigma_i^{a}/\sqrt{N})\right]+\ln\left(1+\prod_{a\in\partial i}\mathcal{G}_{a\rightarrow i}\right)\\
-\left[\sum_{a\in\partial i}
\mathcal{H}_{a\rightarrow i}(+1)+\prod_{a\in\partial i}\mathcal{G}_{a\rightarrow i}\sum_{a\in\partial i}\mathcal{H}_{a\rightarrow i}(-1)\right]/\left(1+\prod_{a\in\partial i}\mathcal{G}_{a\rightarrow i}\right),
\end{split}
\end{equation}
and single example contribution reads
\begin{equation}\label{Sarbm}
\Delta S_a=\ln\cosh(\beta G_a)-\beta^2\Xi_{a}^2/2-\beta G_a\tanh(\beta G_a),
\end{equation}
where we define $\mathcal{H}_{a\rightarrow i}(\xi_i)=\beta^2\Xi_{a\rightarrow i}^2+(\beta G_{a\rightarrow i}+\beta\sigma_i^{a}\xi_i/\sqrt{N})\tanh(\beta G_{a\rightarrow i}+\beta\sigma_i^{a}\xi_i/\sqrt{N})$.

\subsection{Approximate message passing equations}
\label{ampsrbm}
One iteration of the sMP equation (Eq.~(\ref{bp0})) requires the time complexity of the order $\mathcal{O}(MN)$ and memory of the order $\mathcal{O}(MN)$.  The sMP equation can be further simplified by reducing computational complexity. The final equation in physics is called 
Thouless-Anderson-Palmer (TAP) equation~\cite{tap-1977}, and in information theory is named approximate message passing (AMP) equation~\cite{Donoho-2009}.
One strategy is to use large-$N$ limit. We first get
the cavity bias in this limit as $u_{b\rightarrow i}\simeq\frac{\beta\sigma_i^{b}}{\sqrt{N}}\tanh\beta G_{b\rightarrow i}$. Then by applying the same large-$N$ expansion, we obtain
$m_{i\rightarrow a}\simeq m_i-(1-m_i^2)\frac{\beta\sigma_i^a}{\sqrt{N}}\tanh\beta G_{a\rightarrow i}$. Therefore, we get the first AMP equation as follows:
\begin{equation}\label{amp0}
 G_a=\frac{1}{\sqrt{N}}\sum_{i\in\partial a}\sigma_i^{a}m_i-\beta(1-Q)\tanh\beta G_a,
\end{equation}
where $Q\equiv\frac{1}{N}\sum_im_i^2$. Then we define the local field $H_i=\sum_{b\in\partial i}\frac{\sigma_i^b}{\sqrt{N}}\tanh\beta G_{b\rightarrow i}$, and note that $G_{b\rightarrow i}=G_b-\frac{\sigma_i^b}{\sqrt{N}}m_{i\rightarrow b}$,
we can obtain an approximate $H_i$ in the large-$N$ expansion,
\begin{equation}\label{Hi}
 H_i\simeq\sum_{b\in\partial i}\frac{\sigma_i^b}{\sqrt{N}}\tanh\beta G_{b}-\frac{\beta m_i}{N}\sum_{b\in\partial i}(1-\tanh^2\beta G_b).
\end{equation}
The last term in the expression of $H_i$ serves as an Onsager reaction term in a standard TAP equation. Finally we arrive at the second AMP equation:
\begin{equation}\label{amp1}
 m_i\simeq\tanh\left(\sum_{b\in\partial i}\frac{\beta\sigma_i^b}{\sqrt{N}}\tanh\beta G_{b}-\frac{\beta^2 m_i}{N}\sum_{b\in\partial i}(1-\tanh^2\beta G_b)\right).
\end{equation}

Now we have only $N+M$ equations to solve rather than $2NM$ equations in the sMP equation (Eq.~(\ref{bp0})). To make AMP equations converge in a parallel iteration, the time indexes for the variables are important~\cite{Bolt-2014,Lenka-2016}.
Here we write down the closed form of AMP equation with correct time indexes:
\begin{subequations}\label{amprbm}
\begin{align}
G^{t-1}_a&=\frac{1}{\sqrt{N}}\sum_{i\in\partial a}\sigma_i^{a}m_i^{t-1}-\beta(1-Q^{t-1})\tanh\beta G_a^{t-2},\\
m_i^t&\simeq\tanh\left(\sum_{b\in\partial i}\frac{\beta\sigma_i^b}{\sqrt{N}}\tanh\beta G_{b}^{t-1}-\frac{\beta^2 m_i^{t-1}}{N}\sum_{b\in\partial i}(1-\tanh^2\beta G_b^{t-1})\right),
\end{align}
\end{subequations}
where $t$ denotes the time index for iteration. These time indexes just follow the temporal order when we derive the AMP equation from the sMP equation.

\subsection{Statistical analysis of sMP equations and replica computation}
\label{replicasrbm}
Next, we give a statistical analysis of the sMP equation. We first define the cavity field $h_{i\rightarrow a}=\frac{1}{\sqrt{N}}\sum_{b\in\partial i\backslash a}
\sigma_i^b\tanh\beta G_{b\rightarrow i}$. Under the replica symmetric assumption, $h_{i\rightarrow a}$ follows a Gaussian distribution with mean zero and variance $\alpha \hat{Q}$ in the 
large-$N$ limit. We define $\hat{Q}\equiv\left<\tanh^2\beta G_{b\rightarrow i}\right>$. Similarly, $G_{b\rightarrow i}$ also follows a Gaussian distribution with mean zero but
variance $Q$. Therefore, we arrive at the following thermodynamic equation:
\begin{subequations}\label{rbmRM}
\begin{align}
Q&=\int Dz\tanh^2\beta\sqrt{\alpha\hat{Q}}z,\\
\hat{Q}&=\int Dz\tanh^{2}\beta\sqrt{Q}z,
\end{align}
\end{subequations}
where $Dz=\frac{dze^{-z^2/2}}{\sqrt{2\pi}}$. One can expect that when $\alpha$ is small, only one solution of $Q=0$ exists for the above equation, however, at some critical
$\alpha_c$, there is a nontrivial solution of $Q\neq0$, which signals the fixed point of sMP or AMP starts to contain information about the underlying true feature vector.
$\alpha_c$ can be determined by expanding the above equation around $Q=0$. The expansion leads to $\alpha_c=\beta^{-4}$, which implies that when $\alpha<\alpha_c$, $Q=0$ is the
stable solution of the thermodynamic equation, but as long as $\alpha>\alpha_c$, the $Q=0$ is not the stable solution any more. However, as we compare this solution with sMP result on
single instances, $Q\neq0$ solution does not match the numerical simulation very well. To explain this, a replica computation is required. 
%%%%%%%%%%%%%%%%%%%%%%%%%%%%%%%%%%%%%%%%%%%%%%%%%%%%%%%%%%%%%%%%%%%%

Now, we perform a replica computation of the free energy function. Instead of calculating a disorder average of $\ln Z$, the replica trick computes the disorder average of
an integer power of $Z$, then the free energy density (multiplied by $-\beta$) can be obtained as~\cite{Nishimori-2001}
\begin{equation}\label{replica}
 -\beta f=\lim_{n\rightarrow 0,N\rightarrow\infty}\frac{\ln\left<Z^n\right>}{nN},
\end{equation}
where the limit $N\rightarrow\infty$ should be taken first since we can apply the saddle-point analysis~\cite{Nishimori-2001}, and the disorder average is taken over all possible samplings (data) and the random realizations of true feature vector.
The explicit form of $\left<Z^n\right>$ reads
\begin{equation}\label{Zn}
 \left<Z^n\right>=\frac{1}{2^N}\sum_{\{\bs^a,\bx^{{\rm true}}\}}P(\{\bs^a\}|\bx^{{\rm true}})\sum_{\{\bx^\gamma\}}\prod_{a,\gamma}\cosh\left(\frac{\beta\bx^\gamma\bs^a}{\sqrt{N}}\right),
\end{equation}
where $\gamma$ indicates the replica index. We leave the technical details to the appendix~\ref{replica-rbm}, and give the final result here. The free energy function reads,
\begin{equation}\label{freeRBM}
%\begin{align}
 \begin{split}
 -\beta f_{{\rm RS}}=-q\hat{q}+\frac{\hat{r}(r-1)}{2}+\frac{\alpha\beta^2}{2}(1-r)+\int Dz\ln2\cosh(\hat{q}+\sqrt{\hat{r}}z)\\
 +\alpha e^{-\beta^2/2}\int Dy\int Dt\cosh\beta t\ln\cosh\beta
 (qt+\sqrt{r-q^2}y).
 \end{split}
% \end{align}
\end{equation}
and the associated saddle-point equations are expressed as
\begin{subequations}\label{rbmReplica}
\begin{align}
q&=\int Dz\tanh(\hat{q}+\sqrt{\hat{r}}z),\\
r&=\int Dz\tanh^{2}(\hat{q}+\sqrt{\hat{r}}z),\\
\hat{q}&=\alpha\beta^2e^{-\beta^2/2}\int Dt\int Dy\sinh\beta t\tanh\beta(qt+\sqrt{r-q^2}y),\\
\hat{r}&=\alpha\beta^2e^{-\beta^2/2}\int Dt\int Dy\cosh\beta t\tanh^2\beta(qt+\sqrt{r-q^2}y).
\end{align}
\end{subequations}

We make some remarks about the above saddle-point equations. $q$ indicates the typical value of the overlap between the true feature vector and the estimated one, while $r$ indicates the typical value of the overlap between two estimated feature vectors selected from
the posterior probability (Eq.~(\ref{Pobs})). According to the Nishimori condition, $q=r$, implying that the embedded true feature vector follows the same posterior distribution in
Bayesian inference. We verify this point later in numerical solution of the saddle point equations. Assuming $q$ and $r$ are both small values close to zero, Eq.~(\ref{rbmReplica}) in this limit implies that a critical $\alpha_c=\frac{1}{\beta^4}$, above which 
$q=0$ is not a stable solution any more. By expanding Eq.~(\ref{rbmReplica}) around $q=0$ up to the second order $\mathcal{O}(q^2)$, we find
$q\simeq\beta^2(\alpha-\alpha_c)$ when $\alpha$ approaches $\alpha_c$ from above.

Note that by assuming $q=0$ in Eq.~(\ref{rbmReplica}), we obtain Eq.~(\ref{rbmRM}); this is the reason why Eq.~(\ref{rbmRM}) can
predict the correct threshold for transition but could not describe the property of $q\neq0$. This may be because the Gaussian assumption for messages does not generally
hold when the messages start to have partial (even full) alignment with the true feature vector (i.e., $q\neq0$), as also observed in a similar study of the retrieval phase in the Hopfield model by message passing methods~\cite{Mezard-2016}. 
%%%%%%%%%%%%%%%%%%%%%%%%%%%%%%%%%%%%%%%%%%%%%%%%%%%%

The entropy can be derived from the free energy, and the result (Appendix~\ref{replica-rbm}) is given by
\begin{equation}\label{entroRBM}
%\begin{align}
\begin{split}
 s=-2q\hat{q}+r\hat{r}+\frac{\hat{r}(r-1)}{2}-\frac{\alpha\beta^2}{2}(r+1)+\int Dz\ln2\cosh(\hat{q}+\sqrt{\hat{r}}z)\\
 +\alpha e^{-\beta^2/2}\int Dy\int Dt\cosh\beta t\ln\cosh\beta
 (qt+\sqrt{r-q^2}y).
 \end{split}
% \end{align}
\end{equation}

When $q=0$, $s=\ln2-\frac{\alpha\beta^2}{2}$, which coincides with that obtained with cavity method (Appendix~\ref{srbm}). Given $q=0$, that is the data still do not contain information about the hidden feature,
the entropy will become negative once $\alpha>\frac{2\ln2}{\beta^2}\equiv\alpha_{s=0}$. This suggests that the entropy crisis can even occur within $q=0$ regime. Alternatively, at a fixed
$\alpha$, the entropy crisis occurs at a temperature $T_c=\sqrt{\frac{\alpha}{2\ln2}}$. Setting $\alpha_c=\alpha_{s=0}$, one obtains $T_e=\sqrt{2\ln2}$, which distinguishes two cases:
$(i)$ for $T>T_e$, the transition of $q$ from zero to non-zero value takes place after the entropy crisis; $(ii)$ for $T<T_e$, the transition takes place before the crisis. This has clear physical
implications. If the transition occurs before the entropy crisis, the location of transition identified by the replica-symmetric theory is correct. However, the transition
after the crisis is incorrect under the replica-symmetric assumption, because although the replica-symmetric solution is stable, the entropy is negative, violating the fact that
for a system with discrete degrees of freedom, the entropy should be non-negative. According to arguments in Refs.~\cite{Mezard-1984,Gardner-1985}, there exists a discontinuous transition before the entropy crisis takes place, since
the transition can not continuously emerge from a stable replica-symmetric solution. Without further solving complex more-steps replica symmetric breaking equations, we adopt an alternative explanation
of this entropy crisis. Under the current context, if the data size is large enough, the data would shrink the feature space to a sub-exponential regime where the number of
candidate features is not exponential with $N$ any more. In this case, we encounter the entropy crisis. Therefore, the entropy crisis separates an exponential regime from a sub-exponential regime.
The transition for $q$ occurs within the exponential regime if the feature strength is strong enough (large $\beta$).
\begin{figure}
\centering
 \includegraphics[bb=22 19 725 518,scale=0.35]{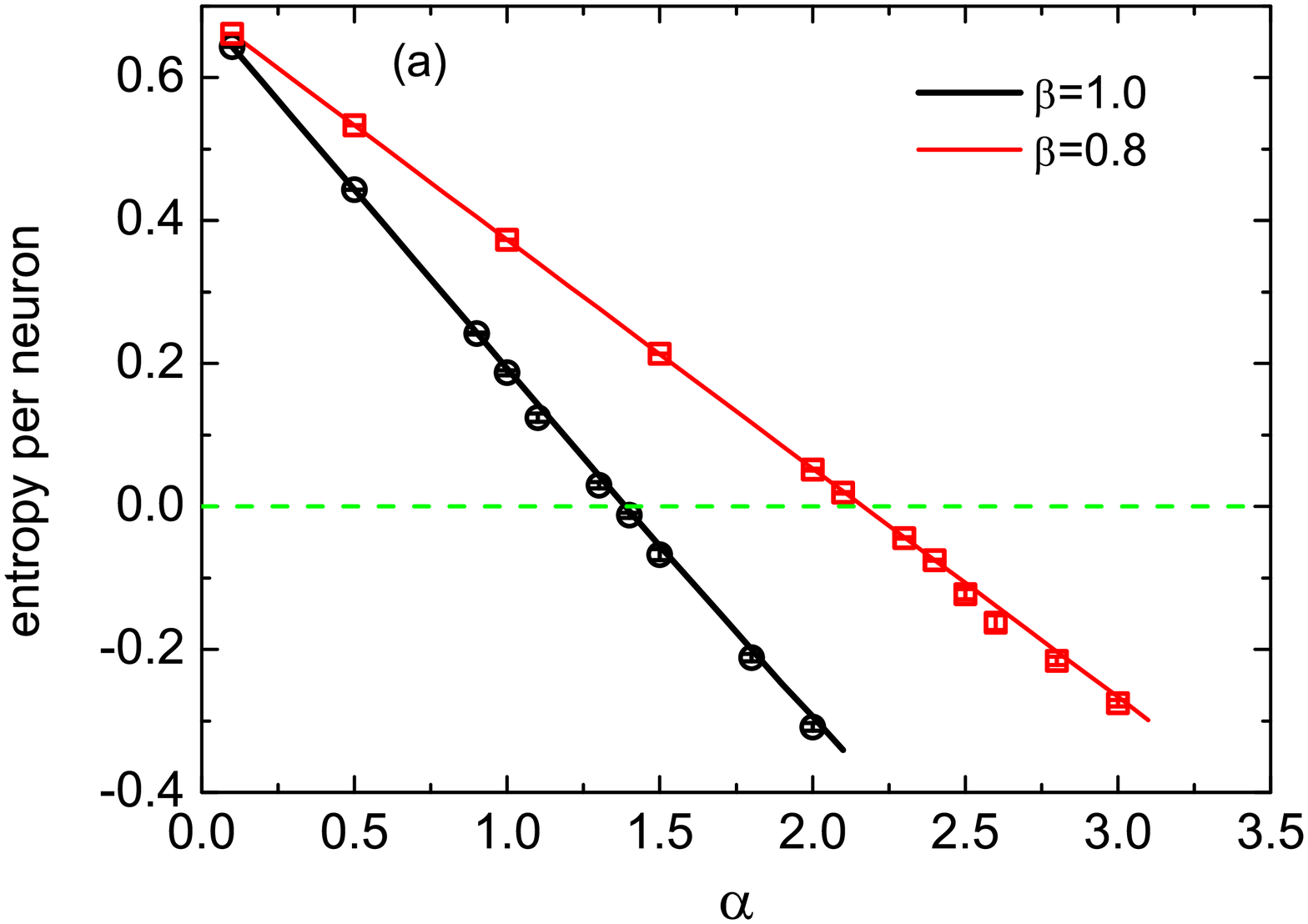}
     \hskip .05cm
  \includegraphics[bb=32 10 728 529,scale=0.35]{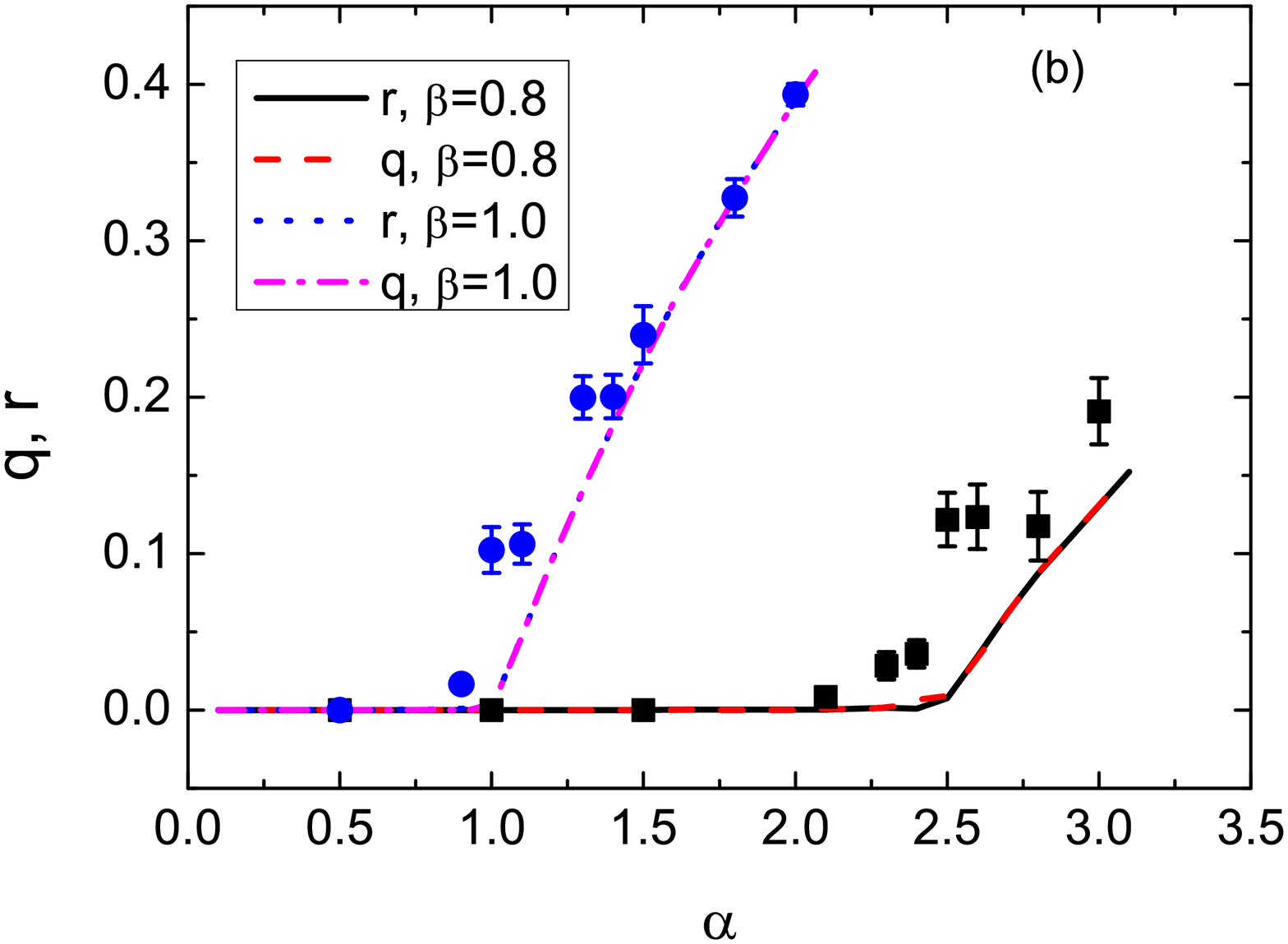}
     \vskip .05cm
  \caption{(Color online) Mean-field result obtained by replica theory compared with simulations carried out by
  running sMP on single instances of RBM. In simulations, we consider $20$ instances of size $N=400$. (a) Entropy per neuron versus
  data density $\alpha$. The lines are replica result, while the symbols are results obtained on single instances. The error bars are
  smaller than the symbol size. (b) Order parameters ($q,r$) versus $\alpha$. $q=r$ as expected. The numerical simulations ($q$) on single 
  instances are indicated by symbols (solid circles for $\beta=1.0$, and solid squares for $\beta=0.8$), while the theoretical predictions of replica computation are indicated by lines.
     }\label{rbm-mf}
 \end{figure}
\subsection{Simulations on single instances compared with theory}
\label{simusrbm}
We use the above mean field theory to analyze single realizations (instances) of the unsupervised learning model. Random
samplings are first generated according to a RBM distribution $P(\bs,h)\propto e^{-\beta E(\bs,h)/\sqrt{N}}$~\cite{Hinton-2006b,Huang-2015b},
where the energy is rescaled by the system size and the inverse temperature, which tunes difficulty level of the learning task. These random samplings
 then serve as the quenched disorder specifying the interaction between the feature vector and the example constraint (Fig.~\ref{rbm}). Finally, by initializing the message
 on each link of the factor graph (Fig.~\ref{rbm}), we run the sMP equation (Eq.~(\ref{bp0})) until it converges within a prefixed precision. From the fixed point, we compute the entropy
 of consistent feature vectors and the overlap between the inferred feature vector and the true one.

We first compare results of the message passing algorithm with those obtained by replica computation. In Fig.~\ref{rbm-mf} (a),
we show the entropy density versus the data density $\alpha$. The entropy characterizes how the number of candidate feature vectors compatible with
the given data changes with the network size $N$. The replica result predicts an entropy crisis, i.e., the entropy becomes negative at some
data size, but the negativity of the entropy is not allowed in a system with discrete degrees of freedom ($\xi_i=\pm1$). This implies that, a discontinuous phase 
transition should be present before the crisis~\cite{Mezard-1984,Gardner-1985}. The results obtained by running sMP coincide perfectly with the
replica result. The entropy density decreases more rapidly with $\alpha$ at larger $\beta$. This is expected, because large $\beta$
indicates strong feature, thus to shrink the feature space to the same size, less data is required compared to the case of detecting weak feature (small $\beta$).

In Fig.~\ref{rbm-mf} (b), we show how order parameters change with $\alpha$. The simulation results agree with the replica
prediction, despite slightly large deviations observed around the transition point. For $\beta>T_e^{-1}$, a first continuous transition occurs at $\alpha_c=1$, which should be correct since
the replica computation is stable and the entropy is positive there. At $\alpha=1$, the system starts to have information about the embedded feature,
and therefore, the overlap $q$ starts to increase even for a finite-size system. The asymptotic behavior of $q$ at a slightly larger $\alpha$ ($>\alpha_c$) is
captured by $\beta^2(\alpha-\alpha_c)$, as already derived in the theory section (Sec.~\ref{replicasrbm}). As predicted by replica computation, at a larger value of
$\alpha=\alpha_{s=0}$, the entropy becomes negative. Equivalently, at this $\alpha$, the entropy vanishes at a critical temperature, and thus
the equilibrium is dominated by a finite number of lowest energy states (so-called condensation phenomenon~\cite{Cavagna-2005,MM-2009}). But the sMP is still
stable, therefore a one-step replica symmetry breaking solution should grow discontinuously from the replica symmetric
solution, and this second discontinuous glass transition is expected before or at the crisis $\alpha$ to resolve the entropy crisis. Intuitively, we expect that
the increasing data will freeze the value of synapses, thus vanishing entropy indicates that the feature space develops isolated
configurations: each configuration forms a single valley in the free energy profile, while the number of these valleys is not exponential any more (but
sub-exponential). To prove this picture, one needs to go beyond the replica symmetric assumption. 

For $\beta<T_e^{-1}$, the continuous
transition takes place after the entropy crisis, which is incorrect. Therefore, according to the above argument,
a discontinuous transition should be expected at or before the crisis data size. Our simulation result also confirms 
the Nishimori condition ($q=r$), that is, when the temperature used to generate data is equal to that used to infer the true feature,
the true feature follows the posterior distribution as well. As a consequence, the overlap between a typical feature configuration and 
the embedded one is equal to the overlap between two typical configurations.
\begin{figure}
\centering  
  \includegraphics[bb=56 18 716 523,scale=0.4]{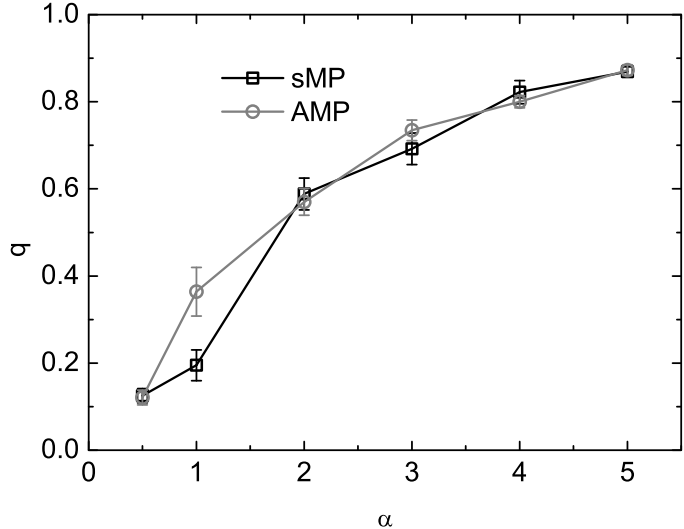}
  \caption{(Color online)  Performance comparison between sMP and AMP for RBM. The model parameters are $N=100$, $\beta=1$. $30$ random instances are 
  considered.
      }\label{amprbp0}
 \end{figure}
 
Secondly, the inference can also be carried out by using AMP with less requirements of computer memory and time. The result is compared with that obtained by sMP, which is shown in Fig.~\ref{amprbp0}.

\begin{figure}
\centering  
  \includegraphics[bb=69 26 719 518,scale=0.4]{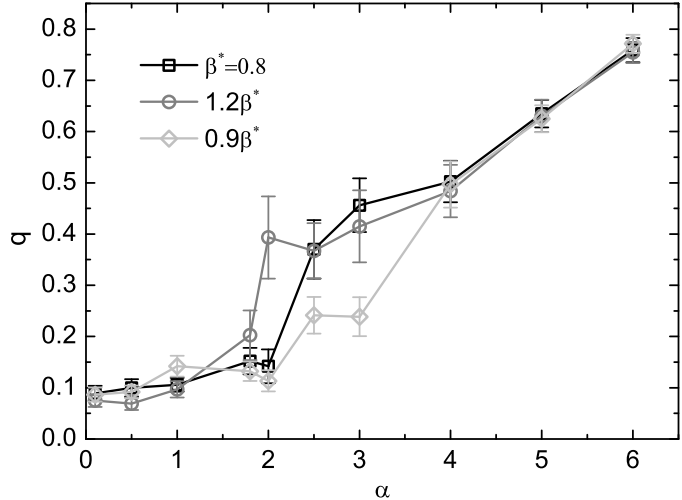}
  \caption{(Color online)  Inference performance with different inference temperatures for RBM. The model parameters are $N=100$, $\beta^{*}=0.8$. $30$ random instances are 
  considered.
      }\label{rbmNopt}
 \end{figure}

We also study the effects of temperature deviation. If the inference is carried out in a different temperature from that used to 
generate the data, is the performance degraded? We address this question by considering two different temperatures: one is slightly larger than
the data temperature ($0.9\beta^{*}$); the other is slightly below the data temperature ($1.2\beta^{*}$), where $\beta^{*}$
denotes the inverse data temperature. As shown in Fig.~\ref{rbmNopt}, when $\beta=\beta^{*}$, the performance is optimal in the inferable regime, compared to other 
inference temperatures, as expected from the Nishimori condition~\cite{Nishimori-2001}. Large fluctuations around the transition point may be caused by finite size effects.

\begin{figure}
\centering  
  \includegraphics[bb=58 16 721 522,scale=0.4]{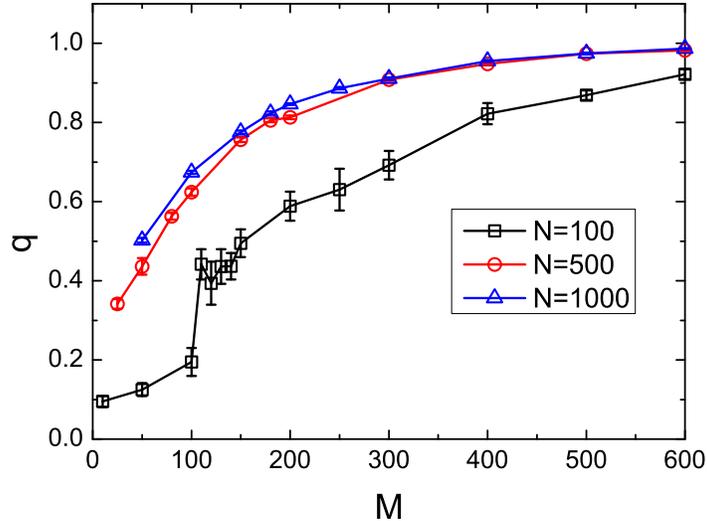}
  \caption{(Color online)  Inference overlap versus the number of examples for RBM with different values of $N$. Feature strength $\beta/\sqrt{N}$
  is kept constant ($0.1$).
      }\label{netsize}
 \end{figure}

Finally, we explore the effect of network size keeping the identical feature strength (Fig.~\ref{netsize}). At a given number of examples, larger network size yields
better performance in terms of prediction overlap. However, the performance seems to get saturated when $N\simeq1000$ at a relatively large $M$ in the current context.
Using a larger network seems to make the unsupervised learning better, but further increasing the network size has a little effect on the performance.

\subsection{Learning temperature parameters from a dataset: how cold is a dataset?}
\label{EM-temp}
In Fig.~\ref{rbmNopt}, we have showed the inference performance with slightly different inference temperatures. Is it possible to infer the true temperatures used to
generate the data itself? If we can learn the temperature parameters, we can know the typical properties of phase transitions intrinsic in the system. This is possible by
applying the Bayesian rule once again. The posterior probability of $\beta$ given the data $\{\bs^{a}\}_{a=1}^{M}$ is given by
\begin{equation}\label{PobsEM}
%\begin{align}
\begin{split}
 P(\beta|\{\bs^a\})&=\sum_{\bx}P(\beta,\bx|\{\bs^{a}\})=\frac{P(\{\bs^a\}|\bx,\beta)P_0(\bx,\beta)}{\int d\beta\sum_{\bx}P(\{\bs^a\}|\bx,\beta)P_0(\bx,\beta)}\\
 &=\frac{1}{Z(\{\bs^{a}\})}\sum_{\bx}e^{-NM\ln\Bigl(2\cosh(\beta/\sqrt{N})\Bigr)}\prod_{a}\cosh\left(\frac{\beta}{\sqrt{N}}\bx^{{\rm T}}\bs^{a}\right)\\
 &\propto e^{-M\frac{\beta^2}{2}}Z(\beta,\{\bs^a\}),
 \end{split}
 %\end{align}
\end{equation}
where we used the uniform prior probability $P_0$ for the hyper-parameters. Note that $Z(\beta,\{\bs^a\})$ is the same partition function as in Eq.~(\ref{Pobs}). We maximize 
the posterior probability with respect to $\beta$, and obtain the self-consistent equation $\beta$ should satisfy:
\begin{equation}
 \label{betaEM}
 \frac{\partial\ln Z(\beta,\{\bs^a\})}{\partial\beta}=N\alpha\beta.
\end{equation}
The left hand side of the above equation is exactly the negative energy ($-N\epsilon$), which can be evaluated by sMP equation (Eq.~(\ref{bp0})). When $N$ is not very large, the equation determining
$\beta$ is given by $\beta=\sqrt{N}\tanh^{-1}\Bigl(-\frac{\epsilon}{\alpha\sqrt{N}}\Bigr)$, from which Eq.~(\ref{betaEM}) is recovered in large $N$ limit. Under the Bethe approximation,
the energy per neuron $\epsilon$ can be computed by $N\epsilon=-\sum_i\Delta\epsilon_i+(N-1)\sum_a\Delta\epsilon_a$, where $\Delta\epsilon_i$ and $\Delta\epsilon_a$ are given respectively by
\begin{subequations}\label{energy}
\begin{align}
\Delta\epsilon_i&=\left[\sum_{a\in\partial i}
\mathcal{H}_{a\rightarrow i}(+1)+\prod_{a\in\partial i}\mathcal{G}_{a\rightarrow i}
\sum_{a\in\partial i}\mathcal{H}_{a\rightarrow i}(-1)\right]/\left(\beta+\beta\prod_{a\in\partial i}\mathcal{G}_{a\rightarrow i}\right),\\
\Delta\epsilon_a&=\beta\Xi_{a}^2+G_a\tanh(\beta G_a).
\end{align}
\end{subequations}

Starting from some initial value of $\beta$, one can iteratively update the value of $\beta$ until convergence within some precision. After one updating, the messages in sMP equation are
also updated. To avoid numerical instability, we used the damping technique, i.e., $\beta(t)=\eta\beta(t)+(1-\eta)\beta(t-1)$, where $t$ denotes the iteration step and $\eta\in[0,1]$ is a damping
factor. It is not guaranteed that there exists unique maximum of the posterior (Eq.~(\ref{PobsEM}))~\cite{Lenka-2016}, but if necessary, one can choose the hyper-parameter 
corresponding to the global maximum of the posterior by running the sMP from different initial conditions.  

In statistics, this iterative scheme is named Expectation-Maximization algorithm~\cite{EM-1977}, where the message updates are called E-step, and the temperature update is called
M-step. In physics, Eq.~(\ref{betaEM}) corresponds to the Nishimori condition ($q=r$, see also Appendix~\ref{replica-rbm} for derivation of the energy function). This means that,
in principle, the hyper-parameter can be learned by iteratively imposing the Nishimori condition~\cite{Mezard-2012,Lenka-2016}. On the Nishimori condition, the state space of the model
is simple~\cite{Nishimori-2001b}, and thus sMP yields informative information about the dominant feature vector. 

\begin{figure}
\centering
 \includegraphics[bb=22 19 725 518,scale=0.3]{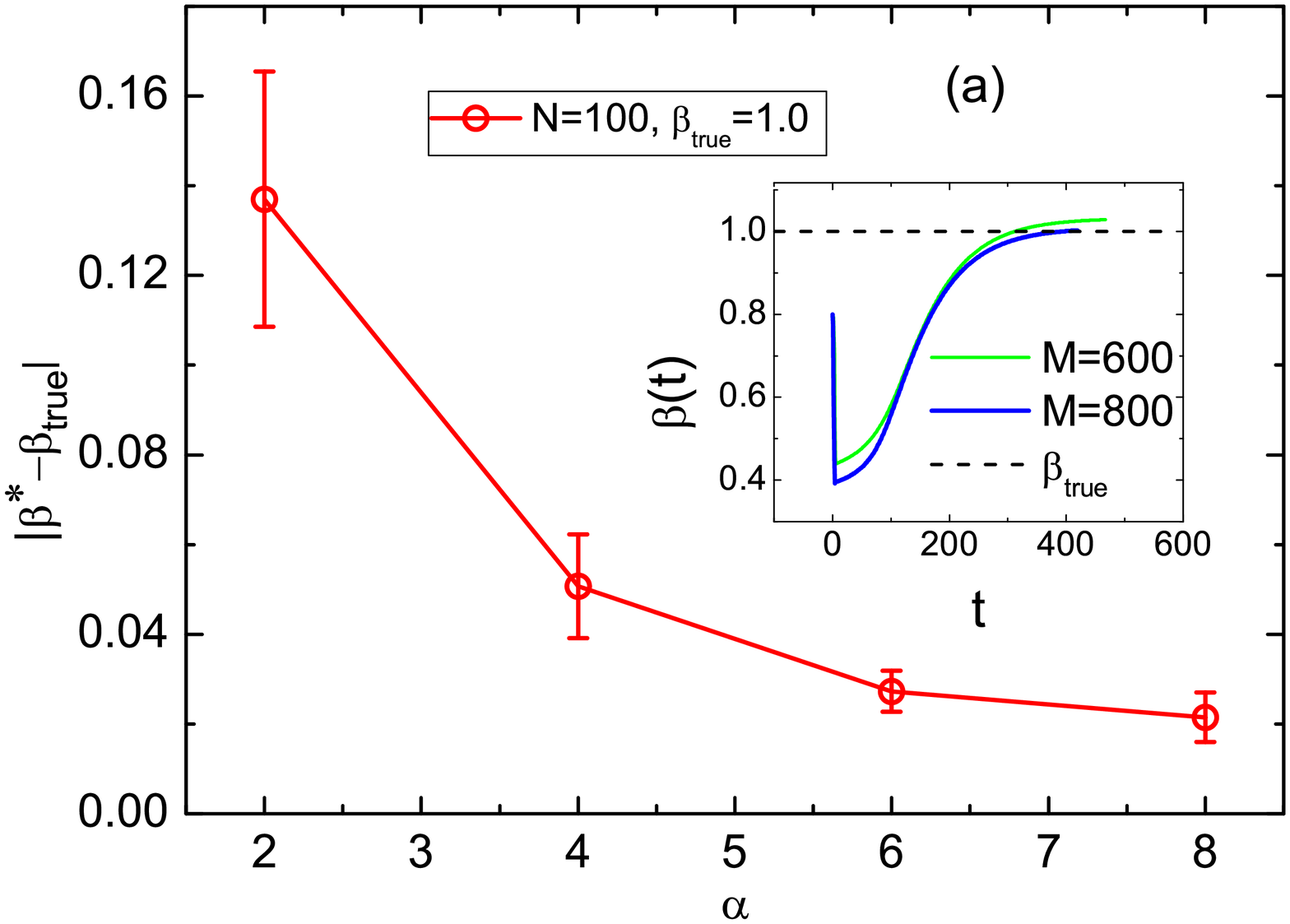}
     \hskip .05cm
  \includegraphics[bb=32 10 728 529,scale=0.3]{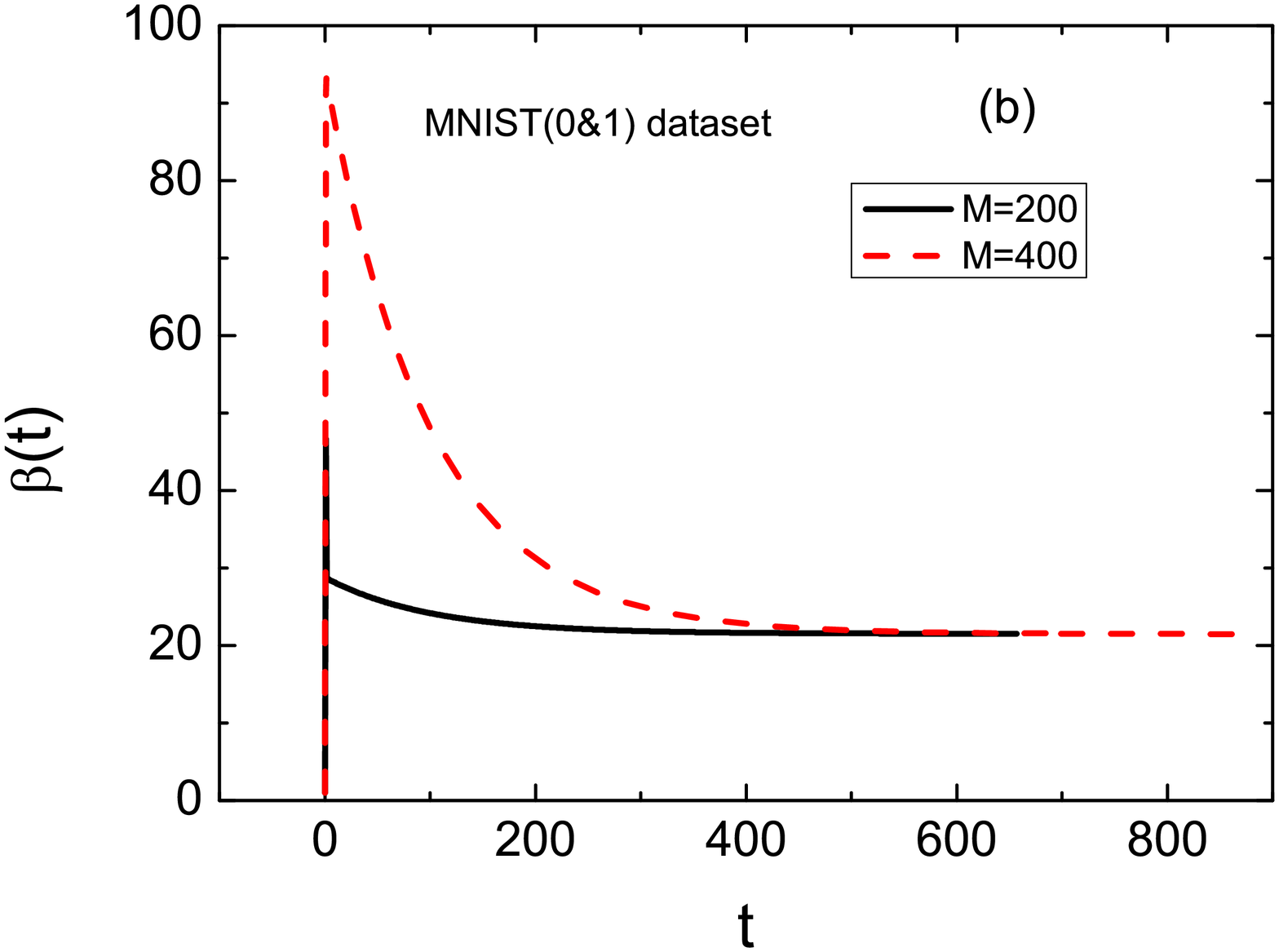}
     \vskip .05cm
 (c)    \includegraphics[bb=0 0 428 308,scale=0.5]{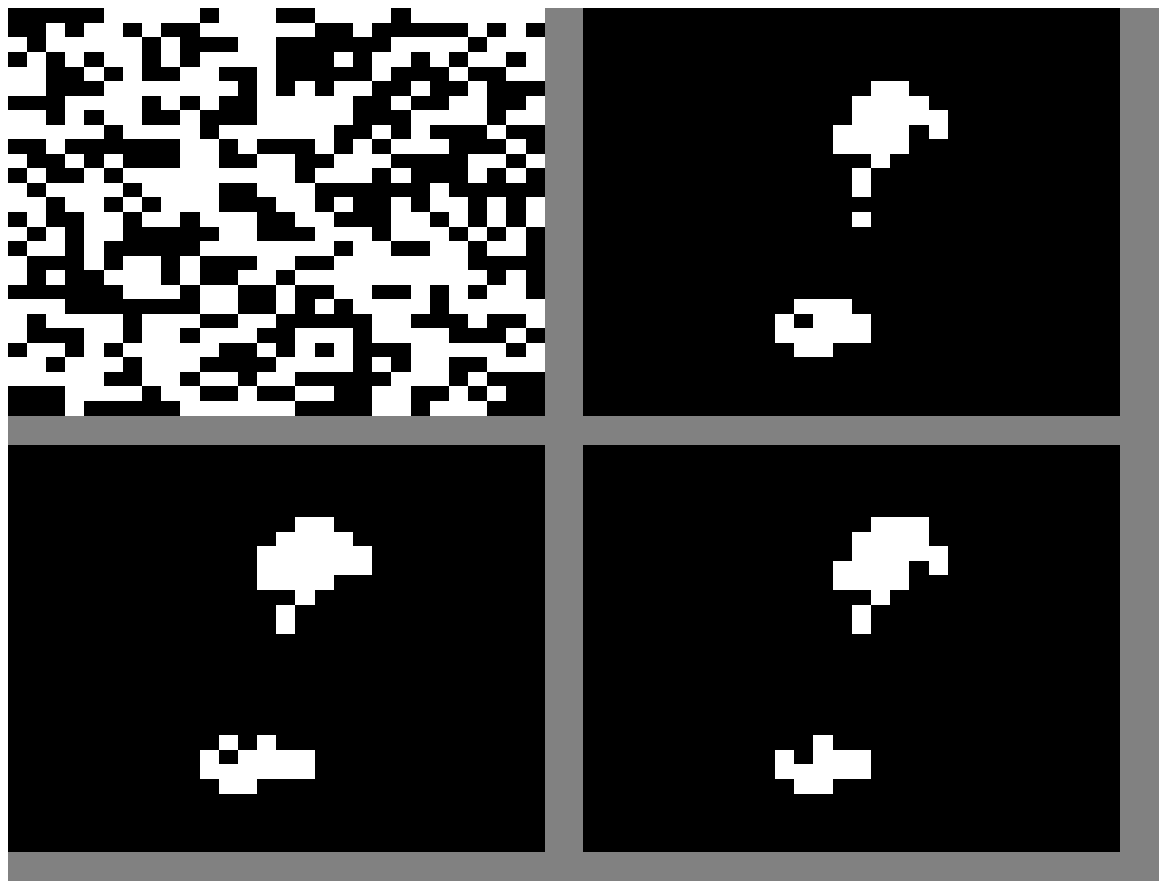}
     \vskip .05cm
  \caption{(Color online) Inference of hyper-parameter $\beta$ in synthetic (RBM) and real (MNIST, digits $0$ and $1$) dataset. (a) Deviation of inferred $\beta$ from the true value decreases with the data size. 
  In simulations, we consider $10$ instances of size $N=100$, and use $\eta=0.02$ and initial value of $\beta_0=0.8$. Two examplar trajectories of $\beta(t)$ are shown in the inset. (b) Examplar trajectories of $\beta(t)$
  are shown for the real dataset. We use $\beta_0=1.0$ and $\eta=0.01$. The fixed point does not change when we use $\beta_0=1.5$. (c) Feature vector ($\bx$) organized as a $28\times28$ matrix.
  They are learned from $M=50$ training images with corresponding $\beta=0.005,0.5,5.0$ and $21.5$ (from the left to the right, and from the top to the bottom). The color black and white indicate active ($\xi_i=+1$) and 
  inactive ($\xi_i=-1$) feature components, respectively.
     }\label{rbm-EM}
 \end{figure}

We know that the temperature parameter is related to the feature strength embedded in the data. Once we learn the temperature, we are able to know how apparent the hidden feature is in
a dataset, and determine the critical data size for unsupervised feature learning. We first test our method in synthetic dataset as already studied in Sec.~\ref{simusrbm}, where the true value of hyper-parameter is known. We then infer
the embedded feature strength in the real dataset (MNIST handwritten digit dataset~\cite{Lecun-1998}), where we do not have any knowledge about the true feature strength. Results are shown in Fig.~\ref{rbm-EM}. For the synthetic data
at $\beta_{\rm true}=1$, as the data size grows, inferred value of $\beta$ gets closer to the true value as expected (Fig.~\ref{rbm-EM} (a)). As shown in the inset, the time (iteration steps) dependent inferred value first
drops to a lower value, and then gradually approaches the true value. With a larger data size (e.g., $M=800$), $\beta$ increases more rapidly after a sudden drop. For the real dataset (Fig.~\ref{rbm-EM} (b)), we observe that the final fixed point of $\beta$
is quite large, implying that the feature strength in the handwritten digits is very strong ($\beta^{*}\simeq21.5$).

We also test effects of $\beta$ on learned features. As shown in the feature map (Fig.~\ref{rbm-EM} (c)), it turns out that the learning works quite well for a broad range of values for $\beta$ except for very small values (e.g., $0.005$), even
when the data is scarce ($M=50$). When $\beta=0.005$, learning fails to identify the meaningful feature. However, a relatively large
value of $\beta$ results in well-structured feature, which has been shown to have the discriminative power for image classification~\cite{Huang-2016}. We conjecture that given the data, these
values of $\beta$ have non-vanishing posterior probability (Eq.~(\ref{PobsEM})). In fact, the value of $\beta=21.5$ has a maximal posterior probability.

\section{The Hopfield model and simplified message passing equations}
\label{ahopf}
\subsection{Message passing equations for an approximate Hopfield model}
\label{mpahopf}
%%%%%%%%%%%%%%%%%%%%%%%%%%%%%%%%%%%%%%%%%%%%%%%%%%%%%%%%%%%%%%%%%%%%%%%
It is interesting to show that one can also perform the same unsupervised learning task by using an associative memory (Hopfield) model defined by
\begin{equation}\label{Hopf}
 \hat{P}(\bx)\propto\prod_{a}e^{\frac{\hb}{2N}\Bigl(\bx^{{\rm T}}\bs^{a}\Bigr)^2},
\end{equation}
where $\hb=\beta^2$. This posterior distribution of feature vectors given the input examples can be obtained by a small-$\beta$ expansion of Eq.~(\ref{Pobs})~\cite{Barra-2012}.
This relationship implies that one can infer the feature vector of a RBM by an approximate Hopfield model, and the feature vector is interpreted as the
stored pattern in the Hopfield model, encoding memory characteristics of the input data. Note that, $\{\bs^a\}$ are still governed by a RBM distribution, whereas, by
applying the associative memory framework, we show many similar interesting properties of the unsupervised learning model.

In analogous to the derivation of Eq.~(\ref{bp0}), we have the sMP corresponding to the posterior probability (Eq.~(\ref{Hopf})):
\begin{equation}\label{bp5}
 m_{i\rightarrow a}=\tanh\left(\frac{\hb}{\sqrt{N}}\sum_{b\in\partial i\backslash a}\sigma_i^{b}\hG_{b\rightarrow i}F_{b\rightarrow i}\right),
\end{equation}
where $\hG_{b\rightarrow i}=\frac{1}{\sqrt{N}}\sum_{k\in\partial b\backslash i}\sigma_k^{b}m_{k\rightarrow b}$, $F_{b\rightarrow i}=1+\frac{\hb C_{b\rightarrow i}}{1-\hb C_{b\rightarrow i}}$ in which
$C_{b\rightarrow i}=\frac{1}{N}\sum_{k\in\partial b\backslash i}(1-m_{k\rightarrow b}^2)$. Details to derive Eq.~(\ref{bp5}) are given in Appendix~\ref{sMP-hopf}.

In the approximate Hopfield model, the Bethe free energy can be constructed similarly, i.e.,  $-\beta Nf_{{\rm RS}}=\sum_{i}\ln Z_i-(N-1)\sum_a\ln Z_a$,
where
\begin{subequations}\label{HopfZ}
\begin{align}
\ln Z_i&=\sum_{a\in\partial i}\left[\frac{\hb}{2}(1/N+\hG^2_{a\rightarrow i})F_{a\rightarrow i}-\frac{1}{2}\ln(1-\hb C_{a\rightarrow i})\right]+\ln2\cosh\hb H_i,\\
\ln Z_a&=\frac{\hb}{2}\hG^2_{a}F_a-\frac{1}{2}\ln(1-\hb C_a),
\end{align}
\end{subequations}
where we define $\hG_{a}=\frac{1}{\sqrt{N}}\sum_{k\in\partial a}\sigma_k^{a}m_{k\rightarrow a}$, $C_a=\frac{1}{N}\sum_{k\in\partial a}(1-m_{k\rightarrow a}^2)$, $F_{a}=1+\frac{\hb C_{a}}{1-\hb C_{a}}$,
and $H_i=\frac{1}{\sqrt{N}}\sum_{b\in\partial i}\sigma_i^{b}\hG_{b\rightarrow i}F_{b\rightarrow i}$.

Similar to the case in RBM, the entropy for the approximate model can be evaluated as
$Ns=\sum_i\Delta S_i-(N-1)\sum_a\Delta S_{a}$, where single feature node contribution reads
\begin{equation}\label{SiHopf}
\begin{split}
\Delta S_i＆=-\sum_{a\in\partial i}\left[\frac{1}{2}\ln(1-\hb C_{a\rightarrow i})+\frac{\hb C_{a\rightarrow i}}{2(1-\hb C_{a\rightarrow i})}
+\frac{\hb}{2}(1/N+\hG^{2}_{a\rightarrow i})F'_{a\rightarrow i}\right]\\
＆+\ln\left(2\cosh(\hb H_i)\right)-(\hb H_i+\hb H'_i)\tanh(\hb H_i),
\end{split}
\end{equation}
and single example contribution reads
\begin{equation}\label{SaHopf}
\Delta S_a=-\frac{1}{2}\ln(1-\hb C_{a})-\frac{\hb C_a}{2(1-\hb C_a)}-\frac{\hb}{2}\hG_a^2F'_a,
\end{equation}
where $F'_{a\rightarrow i}=\frac{\hb C_{a\rightarrow i}}{(1-\hb C_{a\rightarrow i})^2}$,
$F'_{a}=\frac{\hb C_{a}}{(1-\hb C_{a})^2}$, 
and $H'_i=\frac{1}{\sqrt{N}}\sum_{b\in\partial i}\sigma_i^{b}\hG_{b\rightarrow i}F'_{b\rightarrow i}$.

We also derive AMP equations for the Hopfield model. Note that $m_{i\rightarrow a}\simeq m_i-(1-m_i^2)\frac{\hb\sigma_i^a}{\sqrt{N}}\hG_{a\rightarrow i}F_{a\rightarrow i}$.
Therefore $\hG_a\simeq\frac{1}{\sqrt{N}}\sum_{i\in\partial a}\sigma_i^am_i-\hb(1-Q)\hG_a\frac{1}{1-\hb(1-Q)}$, we thus derive the first AMP equation for the Hopfield model as
\begin{equation}\label{amp00}
 \hG_a=\frac{1-\hb(1-Q)}{\sqrt{N}}\sum_{i\in\partial a}\sigma_i^{a}m_i,
\end{equation}
where $Q\equiv\frac{1}{N}\sum_im_i^2$. From the definition of the local field $H_i$, we have $H_i\simeq\frac{1}{1-\hb(1-Q)}\frac{1}{\sqrt{N}}
\sum_{a\in\partial i}\sigma_i^{a}\hG_a-\frac{\alpha}{1-\hb(1-Q)}m_i$ in the large-$N$ expansion. Finally, we derive the second AMP equation:
\begin{equation}\label{amp11}
 m_i\simeq\tanh\left(\frac{\hb}{1-\hb(1-Q)}\frac{1}{\sqrt{N}}
\sum_{a\in\partial i}\sigma_i^{a}\hG_a-\frac{\alpha\hb}{1-\hb(1-Q)}m_i\right).
\end{equation}

Taking the time index into account, the AMP equations can be summarized in a parallel update scheme as follows:
\begin{subequations}\label{amphopf}
\begin{align}
 \hG_a^{t-1}&\simeq\frac{1}{\sqrt{N}}\sum_{i\in\partial a}\sigma_i^am_i^{t-1}-\hb(1-Q^{t-1})\hG_a^{t-2}\frac{1}{1-\hb(1-Q^{t-2})},\\
m_i^{t}&\simeq\tanh\left(\frac{\hb}{1-\hb(1-Q^{t-1})}\frac{1}{\sqrt{N}}
\sum_{a\in\partial i}\sigma_i^{a}\hG_a^{t-1}-\frac{\alpha\hb}{1-\hb(1-Q^{t-1})}m_i^{t-1}\right).
\end{align}
\end{subequations}

\subsection{Thermodynamic equations for the approximate Hopfield model}
\label{replicaahopf}
In this section, we derive the thermodynamic equation. Similarly, the local field defined as $h_{i\rightarrow a}=\frac{1}{\sqrt{N}}\sum_{b\in\partial i\backslash a}\sigma_i^{b}\hG_{b\rightarrow i}F_{b\rightarrow i}$
follows a Gaussian distribution with mean zero and variance $\frac{\alpha Q}{(1-\hb(1-Q))^2}$. The variance can be derived by
noting that $\left<\hG_{b\rightarrow i}^2\right>\simeq Q$. Therefore, we have the following thermodynamic equation for the Hopfield model:
\begin{equation}\label{hopfRM}
 Q=\int Dz\tanh^2\left(\frac{\hb}{1-\hb(1-Q)}\sqrt{\alpha Q}z\right).
\end{equation}
$Q=0$ is a solution of Eq.~(\ref{hopfRM}), however, it is stable only when $\alpha\leq\alpha_c=\left[\frac{1-\hb}{\hb}\right]^2$. This threshold
can be derived by expanding Eq.~(\ref{hopfRM}) around $Q=0$ to the first order. Interestingly, this equation matches the mean-field equation without ferromagnetic part (related to
retrieval phase) derived in
standard Hopfield model~\cite{somp-1985}. Gaussian assumption for messages does not
generally hold, particularly for those messages related to the memorized patterns~\cite{Mezard-2016}.

We also perform replica computation for the approximate Hopfield model. The free energy function is given in the appendix~\ref{replica-hopf}. The associated
saddle-point equations are given as follows:
\begin{subequations}\label{hopfReplica}
\begin{align}
q&=\int Dz\tanh(\hat{q}+\sqrt{\hat{r}}z),\\
r&=\int Dz\tanh^{2}(\hat{q}+\sqrt{\hat{r}}z),\\
\hat{q}&=\frac{\alpha\hb^2q}{1-\hb(1-r)},\\
\hat{r}&=\frac{\alpha\hb^2(\hb q^2+r)}{(1-\hb(1-r))^2}.
\end{align}
\end{subequations}
%%%%%%%%%%%%%%%%%%%%%%%%%%%%%%%%%%%%%%%%%%
Note that, the data is generated by the RBM, but the inference is carried out in an approximate Hopfield model. Although we use the same temperature, the model mismatching 
leads to $q\neq r$. The threshold for the transition of $r$ can be determined by studying the linear stability around $r=0$, and the result is $\alpha^{r}_{c}=\Bigl[\frac{1-\hb}{\hb}\Bigr]^2$, consistent with the 
cavity prediction. When $\alpha$ approaches $\alpha^{r}_c$ from above, $r$ behaves like $r\simeq\frac{\hb^2}{2(1-\hb)}(\alpha-\alpha^r_{c})$.
The transition for $q$ can only be determined numerically, since $r$ could not be assumed a small value, and it follows $r=\int Dz\tanh^2(\sqrt{\hat{r}}z)$ where $\hat{r}=\frac{\alpha\hb^2r}{(1-\hb(1-r))^2}$.
The recursive equation for $r$ in the regime of $q=0$ is exactly the equation derived from the sMP equation (Eq.~(\ref{hopfRM})).

The entropy of the model can also be similarly computed, and reads as follows:
\begin{equation}\label{entrohopf}
 s=\int Dz\ln2\cosh(\hat{q}+\sqrt{\hat{r}}z)-\frac{\alpha}{2}\ln(1-\hb(1-r))-\frac{\alpha\hb((1-\hb(1-r))(1-3r)+2r+2q^2\hb)}{2(1-\hb(1-r))^2}.
\end{equation}

\subsection{Numerical simulations compared with theory}
\label{simuahopf}
\begin{figure}
\centering
    \includegraphics[bb=33 18 714 522,scale=0.35]{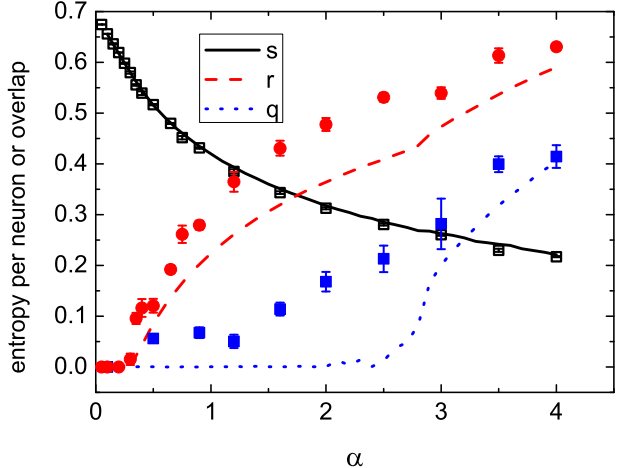}
   \hskip .05cm
  \caption{
  (Color online) Thermodynamic properties of the approximate Hopfield model compared with the inference performed on
  single instances. The lines are replica result, compared with symbols indicating the sMP results. In simulations, we consider $20$ instances of size $N=400$.
  The feature strength $\beta=0.8$.
  }\label{hopf}
\end{figure}

The thermodynamic properties of the approximate Hopfield model are shown in Fig.~\ref{hopf}. First, we show that the entropy
crisis is absent in the Hopfield model, although the inference is carried out by the sMP equation of Hopfield model. This is quite interesting, because within
the associative memory framework, the inference is improved smoothly and there does not exist condensation in the feature
space. Secondly, the replica computation predicts $q\neq r$, as expected from the fact that by applying Hopfield model
approximation, the Nishimori condition does not hold. Thirdly, the simulation results obtained by running sMP equation agree with the theoretical
predictions for the entropy, in spite of observed fluctuations of order parameters caused by finite size effects. The asymptotic behavior of $r$ near $\alpha_c^{r}$ can be analytically
determined by small-$r$ expansion of the saddle-point equation (Eq.~(\ref{hopfReplica})). As already derived in the theory section (Sec.~\ref{replicaahopf}),
$r\simeq\frac{\hb^2}{2(1-\hb)}(\alpha-\alpha^r_{c})$ when $\alpha$ tends to $\alpha_c^{r}$ from above. The transition for $q$ can only be determined by numerically
solving the saddle-point equation. It seems that $q$ changes smoothly to a non-zero value at the same data size as that of the RBM.
Due to the model mismatching, transition for $r$ takes place much more earlier than that for $q$.

\begin{figure}
\centering  
  \includegraphics[bb=46 21 717 521,scale=0.4]{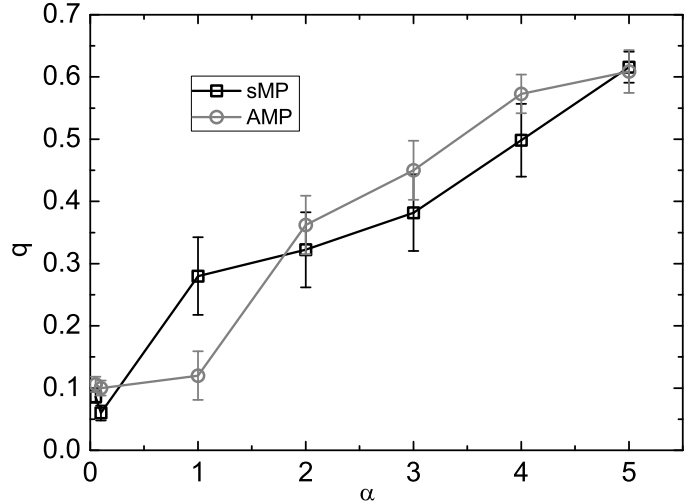}
  \caption{(Color online)  Performance comparison between sMP and AMP for Hopfield model. The model parameters are $N=100,\beta=0.8$. $30$ random instances are 
  considered.
      }\label{amprbp1}
 \end{figure}
 
\begin{figure}
\centering  
  \includegraphics[bb=62 26 718 514,scale=0.4]{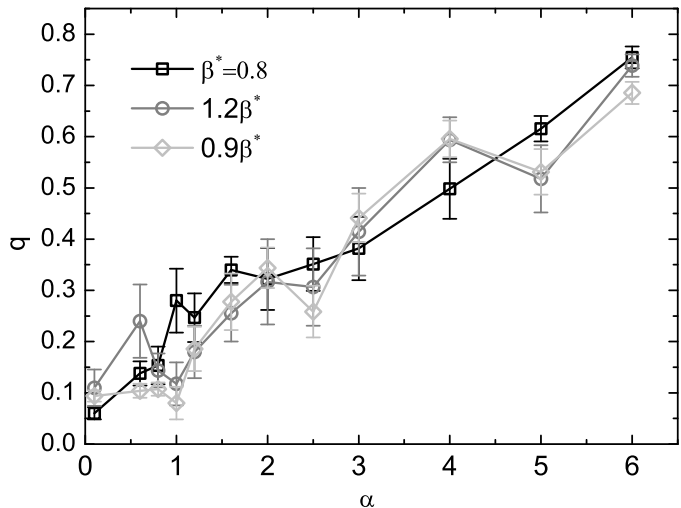}
  \caption{(Color online)  Inference performance with different inference temperatures for Hopfield model. The model parameters are $N=100$, $\beta^{*}=0.8$. $30$ random instances are 
  considered.
      }\label{hopfNopt}
 \end{figure}

Finally, the inference can also be carried out by using AMP with less requirements of memory storage and computer time.
The result is compared with that obtained by sMP, which is shown in Fig.~\ref{amprbp1}. We also study the effects of temperature deviation. As shown in Fig.~\ref{hopfNopt}, even when $\beta=\beta^{*}$, it is not guaranteed that the performance is optimal in the inferable regime, compared to other 
inference temperatures.

\section{Conclusion}
\label{conc}
%%%%%%%%%%%%%%%%%%%%%%%%%%%%%%%%%%%%%%%%%%%%%%%%%%%%%%%%%
In conclusion, we build a physical model of unsupervised learning from a finite number of examples in the framework of RBM. 
Here, we consider binary features rather than real-valued ones; 
this is because binary features are more robust and efficient in large-scale neuromorphic applications~\cite{Bengio-2015}, 
yet it remains open to figure out an efficient algorithm. 
We show that physics method can inspire an efficient (fully-distributed) message passing procedure not only to infer the hidden feature 
embedded in a noisy data, but also to estimate the entropy of candidate features. 
Distinct from conventional slow sampling-based methods, each example in this work
is treated as a constraint on the factor graph, and the message passing carries out a direct Bayesian inference of 
the hidden feature, which marks an important step implementing unsupervised learning in neural networks. In particular, the approximate
message passing equation has low requirements of computer space and time in practical applications.

We show that, the results obtained by the cavity method are consistent with the statistical analysis by replica theory.
The replica theory describes the thermodynamic properties of the unsupervised learning system. It first predicts a discontinuous
phase transition in a restricted Boltzmann machine, signaled by the entropy crisis before the message passing equation loses its stability.
However, if the feature strength is strong enough, there exists another phase transition which is continuous, i.e., the order parameter
(the overlap between the true feature vector and the inferred one) smoothly changes from zero to non-zero value. This continuous transition
will be followed by an additional discontinuous transition at a larger data size. Interestingly, in an approximate Hopfield model,
the entropy crisis is absent, and the entropy decreases much more slowly towards zero. Therefore, there exists a continuous transition from
impossible-to-infer to inferable regime. Unlike the RBM, inference in the Hopfield model does not satisfy the Nishimori condition,
and thus the statistics of metastable states would be very interesting, and its relationship with the dynamics of inference deserves further investigation.

Our work not only derives in detail various kinds of message passing algorithms in a Bayesian framework for practical applications, but also
statistically characterizes the thermodynamic properties of the restricted Boltzmann machine learning with binary synapses, and its
connection with associative memory networks. Many interesting properties related to phase transitions are also revealed. In addition, we derive an iterative equation
to infer the unknown temperature in the data, providing a quantitative measure of how cold a dataset is. This method corresponds to Expectation-Maximization algorithm in statistics~\cite{EM-1977}, and in physics 
iteratively imposing Nishimori condition~\cite{Nishimori-2001,Mezard-2012}. Therefore,
our study forms a theoretical basis of unsupervised feature learning in a single simple RBM, and are expected to be helpful
in constructing a deep architecture for hierarchical information processing, which is currently under way.   

\appendix
\section{Derivation of simplified mesage passing equations for RBM}
\label{sMP-rbm}
We first assume feature components on the factor graph are weakly correlated, then by using the cavity method~\cite{MM-2009}, we define a cavity
probability $P_{i\rightarrow a}(\xi_i)$ of $\xi_i$ on a modified factor graph with example node $a$ removed. Due to the weak correlation assumption, $P_{i\rightarrow a}(\xi_i)$
satisfies a recursive equation (namely belief propagation (BP) in computer science~\cite{Yedidia-2005}):
\begin{subequations}\label{bp00}
\begin{align}
P_{i\rightarrow a}(\xi_{i})&\propto
\prod_{b\in\partial i\backslash
a}\mu_{b\rightarrow i}(\xi_{i}),\label{bp2}\\
\begin{split}
\mu_{b\rightarrow i}(\xi_{i})&=\sum_{\{\xi_{j}|j\in\partial
  b\backslash
  i\}}\cosh\left(\frac{\beta}{\sqrt{N}}\bx^{{\rm T}}\boldsymbol{\sigma}^{b}\right)\prod_{j\in\partial
  b\backslash i}P_{j\rightarrow b}(\xi_{j}),\label{bp1}
\end{split}
\end{align}
\end{subequations}
where the symbol $\propto$ indicates a normalization constant,
$\partial i\backslash a$ defines the neighbors of node $i$ except
constraint $a$, $\partial b\backslash i$ defines the neighbors of
constraint $b$ except visible node $i$, and the auxiliary quantity
$\mu_{b\rightarrow i}(\xi_i)$ represents the contribution from
constraint $b$ to visible node $i$ given the value of
$\xi_i$~\cite{MM-2009}. Eq.~(\ref{bp00}) has been similarly derived to understand RBM in a recent paper~\cite{Huang-2015b}. Here, we exchange the role of
the observed data $\{\bs^{a}\}$ and that of
synaptic interaction (feature vector here), and predict feature vector given the data. Therefore the data (random samplings), rather than
the synaptic interaction in the previous work~\cite{Huang-2015b}, becomes a quenched disorder.

Note that in Eq.~(\ref{bp1}), the sum inside the hyperbolic cosine function with the $i$-dependent term excluded is a random variable following a normal distribution with mean $G_{b\rightarrow i}$ and variance
$\Xi_{b\rightarrow i}^{2}$~\cite{Huang-2015b}, where $G_{b\rightarrow
i}=\frac{1}{\sqrt{N}}\sum_{j\in\partial b\backslash i}\sigma_{j}^{b}m_{j\rightarrow b}$ and
$\Xi^{2}_{b\rightarrow i}\simeq\frac{1}{N}\sum_{j\in\partial b\backslash
i}(1-m_{j\rightarrow b}^{2})$. The cavity magnetization is defined as $m_{j\rightarrow b}=\sum_{\xi_j}\xi_jP_{j\rightarrow b}(\xi_j)$. Thus the intractable sum over
all $\xi_j$ ($j\neq i$) can be replaced by an integral over the normal distribution. Using the magnetization representation~\cite{Huang-2015b},
the BP equation (Eq.~(\ref{bp00})) could be reduced to the simplified message passing equations (see Eq.~(\ref{bp0}) in the main text). 

In physics, the contribution from a single feature node to the partition function, $Z_i$ is obtained via cavity method as $Z_i=\sum_{\xi_i=\pm1}\prod_{b\in\partial i}\mu_{b\rightarrow i}(\xi_i)$; the contribution
of a single data node reads $Z_a=\sum_{\{\xi_j|j\in\partial a\}}\cosh\Bigl(\frac{\beta}{\sqrt{N}}\boldsymbol{\xi}^{{\rm T}}\boldsymbol{\sigma}^a\Bigr)\prod_{j\in\partial a}P_{j\rightarrow a}(\xi_j)$, which can be further
computed by applying the central-limit theorem as well. This calculation is exact only when the underlying factor graph is a tree. However, it is approximately correct when correlations among 
feature components are weak. It needs to be compared with numerical simulations and replica computations.

\section{Simplification of entropy formula for RBM in the limit of $q=0$}
\label{srbm}
First, we compute $\Delta S_i$. By noting that $\Xi^{2}_{a\rightarrow i}=1-\frac{1}{N}$, $\ln\cosh(\beta\sigma_i^a/\sqrt{N})\simeq\frac{\beta^2}{2N}$, and $\beta\sigma_i^a\xi_i/\sqrt{N}\tanh(\beta\sigma_i^a\xi_i/\sqrt{N})\simeq\frac{\beta^2}{N}$,
we have
\begin{equation}
 \Delta S_i=-\sum_{a\in\partial i}\frac{\beta^2}{2}\Xi^2_{a\rightarrow i}-\alpha\beta^2+\frac{\alpha\beta^2}{2}+\ln2,
\end{equation}
where we have used the fact that $G_{a\rightarrow i}=0$ and $\mathcal{G}_{a\rightarrow i}=1$. Analogously, $\Delta S_a$ is simplified to be
\begin{equation}
 \Delta S_a=-\frac{\beta^2}{2}.
\end{equation}
Collecting the above results, we arrive at the final simplified entropy as $s=\ln2-\frac{\alpha\beta^2}{2}$.

\section{Derivation of simplified mesage passing equations for the approximate Hopfield model}
\label{sMP-hopf}
For the approximate Hopfield model, we similarly define the auxiliary quantity
$\mu_{b\rightarrow i}(\xi_i)$ as
\begin{equation}\label{bp0Hopf}
 \begin{split}
  \mu_{b\rightarrow i}(\xi_{i})&=\sum_{\{\xi_{j}|j\in\partial
  b\backslash
  i\}}\exp\left(\frac{\hb}{2N}(\bx^{{\rm T}}\boldsymbol{\sigma}^{b})^2\right)\prod_{j\in\partial
  b\backslash i}P_{j\rightarrow b}(\xi_{j})\\
 &\simeq\int Dz\exp\left(\frac{\hb}{2}\Bigl[\hG_{b\rightarrow i}+\sqrt{C_{b\rightarrow i}}z+\frac{\xi_i\sigma_i^{b}}{\sqrt{N}}\Bigr]^2\right)\\
  &=\frac{1}{\sqrt{1-\hb C_{b\rightarrow i}}}\exp\left(\frac{\hb F_{b\rightarrow i}}{2}\Bigl(\frac{1}{N}+\hG_{b\rightarrow i}^{2}\Bigr)\right)\exp\left(\frac{\hb\hG_{b\rightarrow i}\xi_i\sigma_i^{b}F_{b\rightarrow i}}{\sqrt{N}}\right),
 \end{split}
\end{equation}
where $Dz\equiv\frac{e^{z^2/2}dz}{\sqrt{2\pi}}$. $P_{i\rightarrow a}(\xi_i)$ is the same as that in RBM. Using Eq.~(\ref{bp0Hopf}), the cavity magnetization $m_{i\rightarrow a}$ can thus be derived
as Eq.~(\ref{bp5}).

\section{Replica computation for the RBM model}
\label{replica-rbm}
We first define $u^a=\frac{\bx^{{\rm true}}\bs^a}{\sqrt{N}}$, and $v^{\gamma a}=\frac{\bx^\gamma\bs^a}{\sqrt{N}}$. Both $u^{a}$ and $v^{\gamma a}$ are random variables subject to the covariance structure:
$\left<u\right>=0$, $\left<u^2\right>=1$, $\left<v^\gamma\right>=0$, $\left<(v^\gamma)^2\right>=1$, $\left<uv^\gamma\right>=q^\gamma$, $\left<v^\gamma v^{\gamma'}\right>=r^{\gamma\gamma'}$, where
we have dropped off the data index $a$ because of independence among data samples, and defined the overlap between true feature vector and the estimated one as $q^\gamma=\frac{1}{N}
\sum_i\xi_i^{\gamma}\xi_i^{{\rm true}}$, and the overlap between two estimated feature vectors as $r^{\gamma\gamma'}=\frac{1}{N}\sum_i\xi_i^\gamma\xi_i^{\gamma'}$. Under the replica symmetric assumption,
$q^\gamma=q$ and $r^{\gamma\gamma'}=r$, after introducing the definition of $q^\gamma$ (and $r^{\gamma\gamma'}$) as a delta function, $\left<Z^n\right>$ can be estimated as
\begin{equation}\label{Znrbm}
\begin{split}
\left<Z^n\right>=\int\frac{dqd\hat{q}}{2\pi{\rm i}/N}\int\frac{drd\hat{r}}{2\pi{\rm i}/N}\exp\left[-Nnq\hat{q}-Nr\hat{r}\frac{n(n-1)}{2}-Nn\frac{\hat{r}}{2}+N\ln\int Dz
(2\cosh(\hat{q}+\sqrt{\hat{r}}z))^n\right]\\
\times\exp\left[\alpha N\ln \left\{e^{-\beta^2/2}\int Dy\int Dt\cosh\beta t(\cosh\beta(qt+\sqrt{r-q^2}y))^n\right\}\right],
\end{split}
\end{equation}
where we have written $u=t,v^\gamma=qt+\sqrt{1-r}x^\gamma+\sqrt{r-q^2}y$ ($t,x^\gamma$ and $y$ are standard Gaussian random variables). Finally, we arrived at the following
free energy function:
\begin{equation}\label{frbm}
\begin{split}
 -\beta f_{{\rm RS}}=-q\hat{q}+\frac{\hat{r}(r-1)}{2}+\frac{\alpha\beta^2}{2}(1-r)+\int Dz\ln2\cosh(\hat{q}+\sqrt{\hat{r}}z)\\
 +\alpha e^{-\beta^2/2}\int Dy\int Dt\cosh\beta t\ln\cosh\beta
 (qt+\sqrt{r-q^2}y).
 \end{split}
\end{equation}
The saddle-point equation for the order parameters $\{q,\hat{q},r,\hat{r}\}$ can be derived from $\frac{\partial(-\beta f_{{\rm RS}})}{\partial q}=0$,$\frac{\partial(-\beta f_{{\rm RS}})}{\partial r}=0$,
$\frac{\partial(-\beta f_{{\rm RS}})}{\partial \hat{q}}=0$, and $\frac{\partial(-\beta f_{{\rm RS}})}{\partial \hat{r}}=0$.

Note that to derive the entropy formula, we used $s=(1-\beta\frac{\partial}{\partial\beta})\left<\ln Z\right>(\beta',\beta)|_{\beta'=\beta}$, where $\beta'$ is the inverse temperature
at which the data is generated, and $\beta$ the temperature at which the Bayesian inference is carried out.

\section{Replica computation for the Hopfield model}
\label{replica-hopf}
For the approximate Hopfield model, we replace $\prod_{a,\gamma}\cosh\left(\frac{\beta\bx^\gamma\bs^a}{\sqrt{N}}\right)$ with 
$\prod_{a,\gamma}e^{\frac{\hb}{2}\left(\frac{\bx^\gamma\bs^a}{\sqrt{N}}\right)^2}$ in Eq.~(\ref{Zn}). The subsequent calculation proceeds similarly to the appendix~\ref{replica-rbm}.
Analogously, we have
\begin{equation}\label{Znhopf}
\begin{split}
\left<Z^n\right>=\int\frac{dqd\hat{q}}{2\pi{\rm i}/N}\int\frac{drd\hat{r}}{2\pi{\rm i}/N}\exp\left[-Nnq\hat{q}-Nr\hat{r}\frac{n(n-1)}{2}-Nn\frac{\hat{r}}{2}+N\ln\int Dz
(2\cosh(\hat{q}+\sqrt{\hat{r}}z))^n\right]\\
\times\exp\left[\alpha N\ln \left\{e^{-\beta^2/2}\int Dy\int Dt\cosh\beta t\Bigl(\frac{1}{\sqrt{1-\hb(1-r)}}e^{\frac{\hb(qt+\sqrt{r-q^2}y)^2}{2(1-\hb(1-r))}}\Bigr)^n\right\}\right].
\end{split}
\end{equation}
Using the replica trick defined in Eq.~(\ref{replica}), we obtain the free energy function
\begin{equation}\label{fhopf}
 -\hb f_{{\rm RS}}=-q\hat{q}+\frac{\hat{r}}{2}(r-1)+\int Dz\ln2\cosh(\hat{q}+\sqrt{\hat{r}}z)-\frac{\alpha}{2}\ln(1-\hb(1-r))+\frac{\alpha\hb(r+\hb q^2)}{2(1-\hb(1-r))},
\end{equation}
where we used the identity $\int Dy\int Dt[qt+\sqrt{r-q^2}y]^2\cosh\beta t=e^{\hb/2}(\hb q^2+r)$. The saddle-point equations can be derived similarly.

\section{Generalization to the case of the hidden neuron with an external field in RBM}
\label{gener}
The sMP for RBM can be easily generalized to take into account external fields of visible neurons and the hidden neuron. Here, for simplicity, we consider only the case of
hidden neuron with external field. The external field has binary values as well, defined by $B_h\phi_0$ ($B_h=\pm1$). The only modification to the factor graph in Fig.~\ref{rbm} is to add 
one additional variable node named by $\lambda$ for the unknown external field. The additional node $\lambda$ is connected to all data nodes. Following the similar procedure as in Appendix~\ref{sMP-rbm},
we obtain the following four kinds of messages:
\begin{subequations}\label{rbmext}
\begin{align}
m_{\lambda\rightarrow a}&=\tanh\left(\sum_{b\in\partial \lambda\backslash
a}u_{b\rightarrow \lambda}\right),\\
u_{b\rightarrow \lambda}&=\tanh^{-1}\left(\tanh(\beta G_{b\rightarrow \lambda})\tanh(\beta\phi_0)\right),\\
m_{i\rightarrow a}&=\tanh\left(\sum_{b\in\partial i\backslash
a}u_{b\rightarrow i}\right),\\
u_{b\rightarrow i}&=\frac{1}{2}\ln\frac{\frac{1+m_{\lambda\rightarrow b}}{2}\cosh\beta(G_{b\rightarrow i}+\sigma_i^b/\sqrt{N}+\phi_0)+\frac{1-m_{\lambda\rightarrow b}}{2}\cosh\beta(G_{b\rightarrow i}+\sigma_i^{b}/\sqrt{N}-\phi_0)}
{\frac{1+m_{\lambda\rightarrow b}}{2}\cosh\beta(G_{b\rightarrow i}-\sigma_i^b/\sqrt{N}+\phi_0)+\frac{1-m_{\lambda\rightarrow b}}{2}\cosh\beta(G_{b\rightarrow i}-\sigma_i^{b}/\sqrt{N}-\phi_0)}.
\end{align}
\end{subequations}
Once $\phi_0=0$, the original sMP for the RBM in the main text is recovered. To derive sMP for the case of visible neurons with external fields, an additional central limit theorem applies to
the interacting external fields ($\{B_i\phi_0\}$), which leads to introducing a joint cavity probability $P_{i\rightarrow a}(\xi_i,B_i)$ as well as $\mu_{b\rightarrow i}(\xi_i,B_i)$ where $B_i=\pm1$.
%%%%%%%%%%%%%%%%%%%%%%%%%%%%%%%%%%%%%%%%%%%%%%%%%%%%%%%
%\section*{Acknowledgments}

\begin{acknowledgments}
I am very grateful to Taro Toyoizumi, Lukasz Kusmierz, Alireza Goudarzi and Roberto Legaspi for attending a series of lectures about this work and their useful feedback. 
I thank Lukasz Kusmierz for a careful reading of the manuscript and his useful feedback.
This work was supported by the program for Brain Mapping by Integrated Neurotechnologies
for Disease Studies (Brain/MINDS) from Japan Agency for Medical Research and development, AMED.
\end{acknowledgments}
%%%%%%%%%%%%%%%%%%%%%%%%%%%%%%%%%%%%%%%%%%%%%%%%%%%%%%%%%%%%%%%
%\bibliography{ref}

%%%%%%%%%%%%%%%%%%%%%%%%%%%%%%%%%%%%%%%%%%%%%%%%%%%%%%%%%%%%%%%%%%%%%

\end{document}